%% file: main.tex
\pdfoutput=1
\def\paperrelease{1}

\documentclass[letterpaper, 10pt, conference]{ieeeconf}

\IEEEoverridecommandlockouts

\usepackage{graphicx}
\usepackage{amsmath}
\usepackage{amssymb}
\usepackage[table,xcdraw]{xcolor}
\usepackage{subfigure}

\usepackage{tabularx}
\usepackage{makecell}
\usepackage{multirow}
\usepackage{booktabs}
\usepackage{array}
\usepackage{threeparttable}

\usepackage{cite}
\makeatletter
\let\NAT@parse\undefined
\makeatother
\usepackage[numbers]{natbib}

\usepackage{lipsum}
\usepackage[font=small,labelfont=bf,labelsep=period]{caption}
\usepackage{cuted}
\usepackage{capt-of}
\usepackage{siunitx}
\let\labelindent\relax
\usepackage{enumitem}
\usepackage{afterpage}
\usepackage{aprl_acronyms}
\usepackage{aprl_misc}

\usepackage[colorlinks=true, citecolor=blue, linkcolor=cyan, urlcolor=magenta]{hyperref}

\usepackage{svg}
\usepackage[percent]{overpic}
\usepackage{dblfloatfix}
\usepackage{subcaption}

\usepackage{cuted}
\usepackage{caption}

\title{\LARGE \bf MR.ScaleMaster: Scale-Consistent Collaborative Mapping \\ from Crowd-Sourced Monocular Videos}

\ifnum\paperrelease=1
    \input{authors.tex}
\else
    \author{Anonymous Authors}
\fi

\IEEEaftertitletext{\vspace{-17pt}}

\begin{document}
\maketitle
\input{src_v3/figures/fig_hero_shot}
\thispagestyle{empty}
\pagestyle{empty}

\input{src_v3/abstract}
\input{src_v3/introduction}
\input{src_v3/related_works}
\input{src_v3/system_overview}

\input{src_v3/method.tex}

\input{src_v3/experiments}
\input{src_v3/limitation}
\input{src_v3/conclusion}

\bibliographystyle{IEEEtran}
{\footnotesize
\bibliography{ref}
}

\end{document}

%% file: authors.tex
\author{Hyoseok Ju$^{1}$ and Giseop Kim$^{1*}$%
\thanks{$^{1}$Department of Robotics and Mechatronics Engineering, DGIST,
Daegu, Republic of Korea.
\texttt{\{hyoseokju,gsk\}@dgist.ac.kr}. $^{*}$Corresponding author.}}

%% file: src_v3/figures/fig_hero_shot.tex
\begin{strip}
    \centering
    \includegraphics[width=0.90\textwidth]{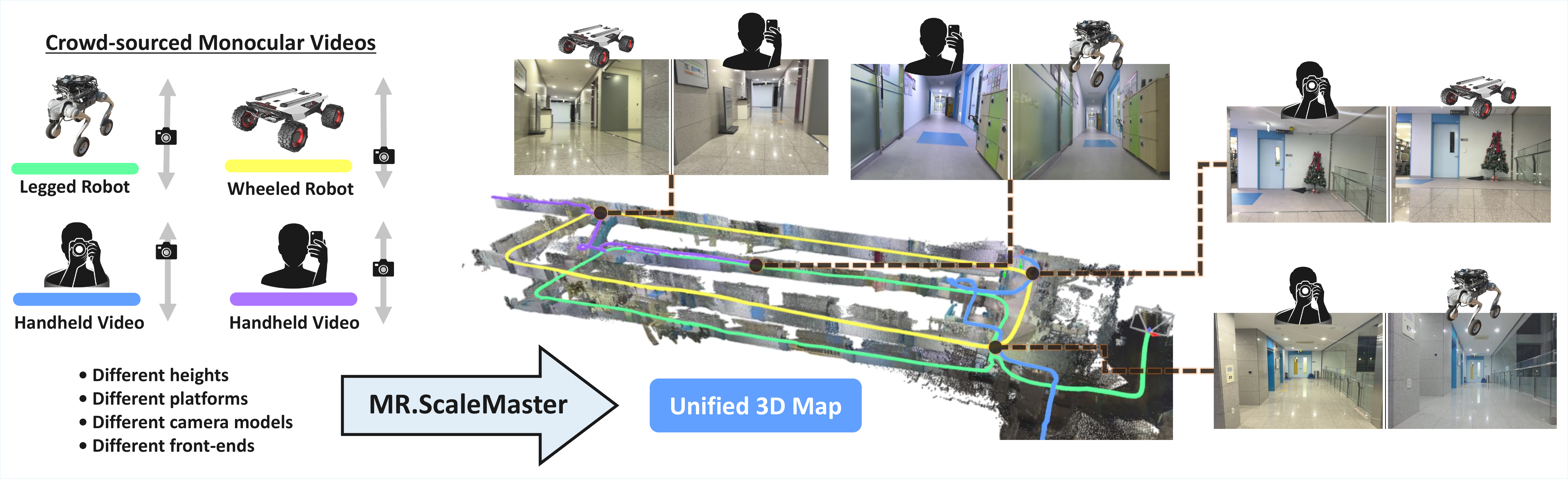}
    \captionof{figure}{Real-world heterogeneous multi-robot mapping with MR.ScaleMaster in a multi-floor indoor environment. Four agents, a legged robot (\textcolor{green!70!black}{\textbf{green}}), a wheeled robot (\textcolor{yellow!80!black}{\textbf{yellow}}), and two handheld cameras operated by different users (\textcolor{blue}{\textbf{blue}}, \textcolor{violet}{\textbf{purple}}), collaboratively build a unified dense 3D map. Inset pairs show inter-agent loop closures across different platforms.}
    \label{fig:hero_shot}
    \vspace{-5mm}
\end{strip}

%% file: src_v3/abstract.tex
\begin{abstract}
Crowd-sourced cooperative mapping combines monocular sessions from different
front-ends, each an independently reconstructed keyframe sequence. Each
session has its own local coordinate frame and may follow an incompatible
scale convention. We present MR.ScaleMaster, a backend that accepts image,
$\mathrm{Sim}(3)$ pose, and point-map packets without requiring a common
reconstruction model, camera intrinsics, or front-end-provided metric scale.
Our \acf{CFL} uses a shared matcher for image
correspondences but retrieves matched 3D points from the input point maps,
so its scale estimates the inter-session ratio. \acf{SPC} initializes
session scales for a $\mathrm{Sim}(3)$ anchor-node graph, which
then corrects residual scale and drift. For
front-ends exposing an incremental scale trajectory, an agent-side \acf{SCA}
rejects or rolls back false intra-session loops that would otherwise
collapse the session scale. We evaluate seven front-ends on KITTI and five on
CODa. Based on ground-truth path lengths, session scales in our heterogeneous
KITTI setting differ by $57$--$103\times$. \acs{CFL} reduces mean ATE by 37\%
and inter-session scale error by 60\% relative to loop factors from independent
pairwise reconstructions. Adding \acs{SPC} raises the reductions to 74\% and
77\%, respectively. On CODa, five-session fusion improves
mean ATE over single-session runs for all five front-ends, while
joint fusion registers 15 sessions from three different front-ends in a
single map. Code will be released.
\end{abstract}

%% file: src_v3/introduction.tex
\section{INTRODUCTION}
\label{sec:intro}

Crowd-sourced cooperative mapping turns cameras on diverse platforms into a
shared map~\cite{sarlin2022lamar}. It must fuse partial trajectories produced
by different visual \ac{SLAM} or reconstruction
pipelines~\cite{slamhandbook}.

Metric sensing makes scale observable but requires additional hardware and
calibration~\cite{patel2023covins,tian2022kimeramulti,
lajoie2024swarmslam}. Monocular front-ends avoid such sensing, but each
independently reconstructed keyframe sequence has an unknown scale, preventing
direct fusion under scale drift or incompatible conventions. We refer to each
such sequence as a session; it has its own local coordinate frame, and one
front-end may contribute multiple sessions.

Recent 3D vision foundation models~\cite{leroy2024mast3r,wang2025vggt,
wang2025pi3,wang2025cut3r,chen2026ttt3r,zhang2026loger,chen2026lingbot}
provide dense geometry without known intrinsics; subsequent systems extend
them to long sequences~\cite{deng2025vggt,lee2026swiftvggt,zhang2026vggt}
and shared-pipeline multi-agent mapping~\cite{zhou2025multi,cao2026como3r,gong2026magist3r}.
Their predictions, however, carry no consistent scale. The ScaleMaster benchmark~\cite{ju2026have} documented both
intra-session scale drift and inter-session scale discrepancies in
feed-forward monocular SLAM. It did not, however, address how such
discrepancies should be resolved when sessions are produced by independently
selected front-ends. In crowd-sourced mapping, session poses
and point maps may follow different scale conventions, while the pairwise
geometry used for cross-session matching may introduce another.
To our knowledge, no prior system has fused monocular sessions from
independently selected front-ends with incompatible scale conventions.
This heterogeneous setting therefore poses two challenges. First,
cross-session loop measurements must be constructed without importing the
matcher's independent scale convention (\figref{fig:motivation}). Second,
intra-session scale drift and unknown inter-session scale ratios must be
aligned in a common frame.

\afterpage{\input{src_v3/figures/fig_motivation}}

We propose \textbf{MR.ScaleMaster} (Multi-Robot ScaleMaster), a cooperative
mapping backend for heterogeneous monocular front-ends. It constructs
cross-session measurements using the \acf{CFL}. \acf{SPC} initializes
session scales for a $\mathrm{Sim}(3)$ anchor-node graph, which
corrects residual scale and drift. For front-ends that expose an incremental
scale trajectory, an agent-side \acf{SCA}
(\secref{sec:scale_alarm}) additionally rejects or rolls back
false intra-session loop closures that would otherwise collapse a session's
scale (\figref{fig:scale_alarm}).

This paper makes three contributions:
\begin{enumerate}
    \item \textbf{A front-end-agnostic cooperative backend}
    (\secref{sec:system}): sessions enter as (image, $\mathrm{Sim}(3)$ pose,
    point map) packets, and the server requires no knowledge of the front-end
    model or camera intrinsics, nor any front-end-provided metric scale.

    \item \textbf{Scale-consistent fusion across incompatible conventions}
    (\secref{sec:self_calibration}--\ref{sec:sim3_anchor}): \acs{CFL}s
    built from the input packet geometry remove the matcher's scale from
    cross-session measurements; \acs{SPC} initializes session scales,
    and a $\mathrm{Sim}(3)$ anchor-node graph corrects residual scale and drift
    in one frame.

    \item \textbf{A comprehensive evaluation} (\secref{sec:exp}): seven
    monocular front-ends on KITTI, five also evaluated on CODa, including
    scale ratios up to $103\times$, paired backend comparisons, and joint
    fusion of three different front-ends.
\end{enumerate}

%% file: src_v3/figures/fig_motivation.tex
\begin{figure}[t]
    \centering
    \includegraphics[width=\columnwidth]{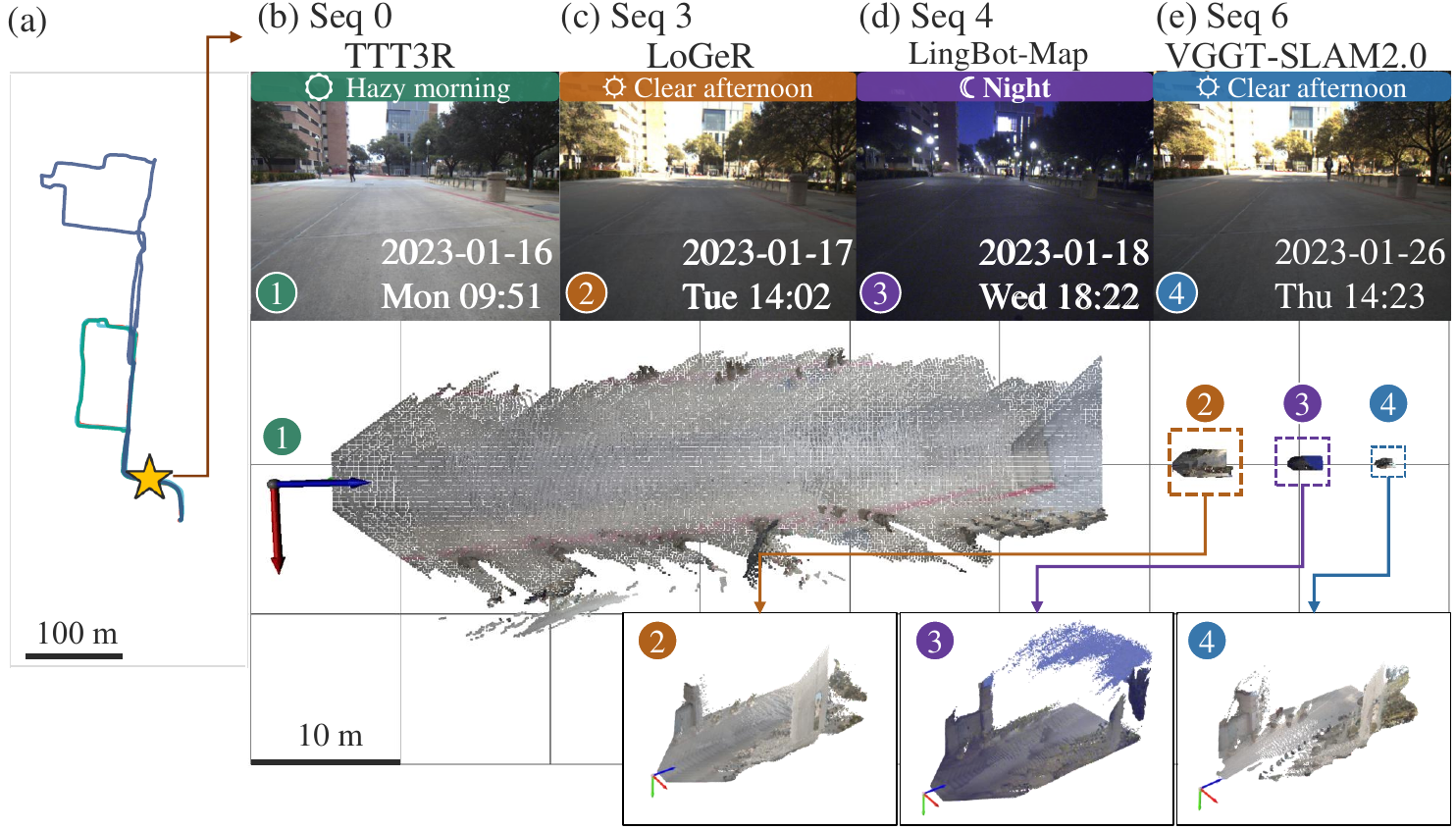}
    \caption{Crowd-sourced CODa inputs. (a) One location (star) revisited
    across four sequences, days, and lighting conditions; (b)--(e) query views
    of the four visits, each reconstructed by a different front-end (below).
    Bottom: raw point maps plotted together in native units without
    cross-front-end rescaling; TTT3R (1) spans tens of units, whereas maps
    2--4 (enlarged) span only a few. After ground-truth alignment, meters per
    unit differ by up to $41\times$.}
    \label{fig:motivation}
    \vspace{-5mm}
\end{figure}

%% file: src_v3/related_works.tex
\section{Related Work}
\label{sec:rel}

\input{src_v3/figures/fig_system_overview}

\subsection{Multi-Robot Cooperative Mapping}

COVINS-G~\cite{patel2023covins}, Kimera-Multi~\cite{tian2022kimeramulti}, and
Swarm-SLAM~\cite{lajoie2024swarmslam} fuse metric local estimates through
centralized, distributed, or decentralized optimization; COVINS-G accepts
arbitrary visual--inertial front-ends, whose estimates are metric.
Lajoie et al.~\cite{lajoie2025foundation} estimate inter-robot loops with 3D
foundation models and optimize their scale in the pose graph, but assume
metric local odometry. CCM-SLAM~\cite{schmuck2019ccm} and
ORB-SLAM3~\cite{campos2021orbslam3} merge monocular maps through
$\mathrm{Sim}(3)$ alignment within one keypoint-based pipeline. Anchor-node formulations retain local pose graphs
and register them in a shared $\mathrm{SE}(3)$
frame~\cite{kim2010multiple,mcdonald20116}; we extend this construction to per-session $\mathrm{Sim}(3)$ anchors, so
that each anchor carries the session's unknown scale and every inter-session
loop factor constrains four vertices instead of two.

Recent monocular collaborative systems include Multi-Agent Monocular Dense
SLAM~\cite{zhou2025multi}, which extends MASt3R-SLAM with loop-based fusion,
CoMo3R-SLAM~\cite{cao2026como3r}, which combines learned point-map
matching, $\mathrm{Sim}(3)$ gauge synchronization, and joint refinement, and
MAGiSt3R~\cite{gong2026magist3r}, which merges VGGT submaps with a learned
module operating on backbone tokens. All three organize collaboration around
a common reconstruction pipeline; MAGiSt3R also evaluates an alternative
backbone, replacing it for all agents.
MR.ScaleMaster instead accepts session-local packets from independently
selected front-ends without assuming a common model or shared scale
convention. Its depth-prior-based \acs{SPC} (\secref{sec:self_calibration})
provides only an initialization for optimization, not a common metric gauge.

\tabref{tab:design_scope} summarizes the agent-side reconstruction assumptions
and scale treatment of these system classes.

\input{src_v3/tables/tab_design_scope}

\subsection{Monocular SLAM and 3D Reconstruction Priors}

Classical monocular SLAM handles scale drift through $\mathrm{Sim}(3)$
loop closure and global optimization~\cite{strasdat2010rss,mur2015orb}.
MASt3R-SLAM~\cite{murai2025mast3r} builds on pairwise MASt3R~\cite{leroy2024mast3r} point maps, while VGGT-SLAM~2.0~\cite{maggio2025vggt-slam2} redesigns its
factor graph around learned reconstruction ambiguities.
MR.ScaleMaster uses MASt3R only for retrieval and correspondences, not for
loop-factor geometry (\secref{sec:loop_factor}).

\subsection{Robust Multi-Robot Pose-Graph Optimization}

Robust pose-graph methods suppress inconsistent measurements using switchable
constraints~\cite{sunderhauf2012switchable}, dynamic covariance
scaling~\cite{agarwal2013robust}, mixture models~\cite{olson2013inference}, or
pairwise consistency tests~\cite{mangelson2018pairwise}. In monocular
$\mathrm{Sim}(3)$ graphs, however, an erroneous loop may be absorbed through
scale distortion while retaining a small residual. The Scale Collapse
Alarm (\secref{sec:scale_alarm}) targets this failure mode on the agent side.

%% file: src_v3/figures/fig_system_overview.tex
\begin{figure*}[!t]
    \centering
    \includegraphics[width=1.0\textwidth]{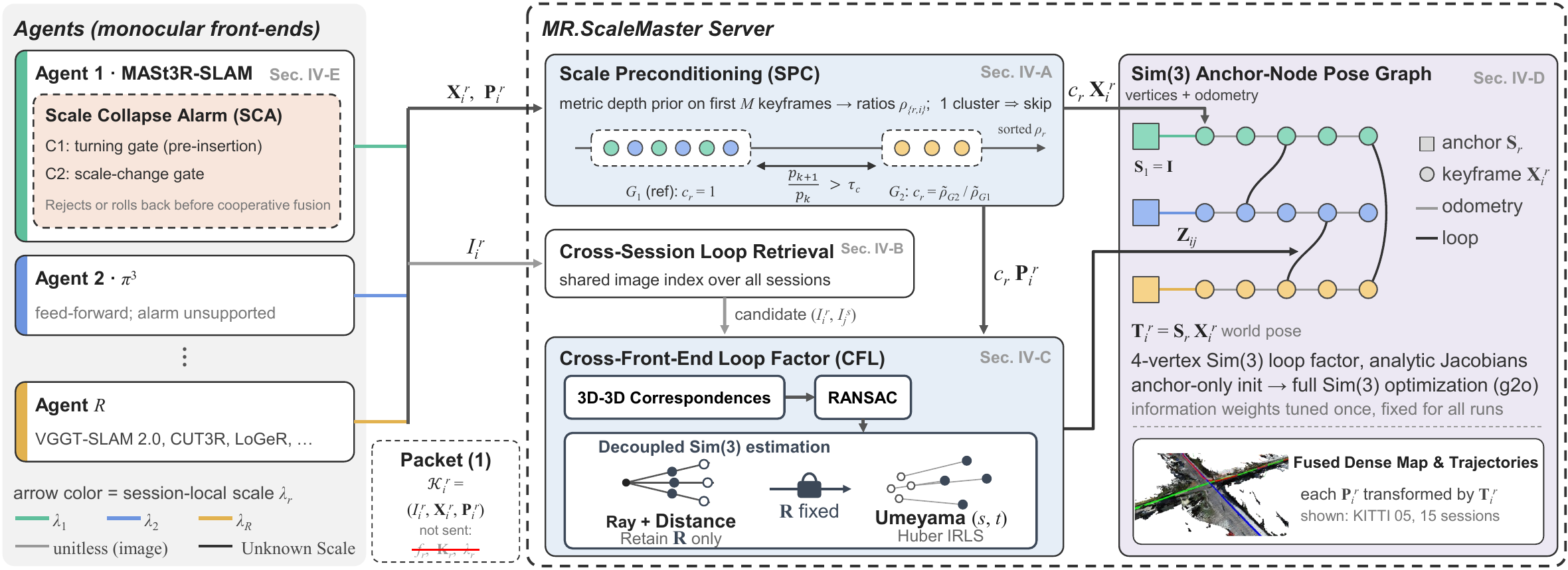}
    \caption{System overview of MR.ScaleMaster. Agent-side \acs{SCA} rejects or
    rolls back false intra-session loops. The server applies \acs{SPC} to
    initialize session scales, establishes cross-session correspondences,
    retrieves matched 3D points from the input point maps, constructs
    $\mathrm{Sim}(3)$ \acs{CFL}s, and optimizes them with odometry in an
    anchor-node graph.}
    \vspace{-5mm}
    \label{fig:system_overview}
\end{figure*}

%% file: src_v3/tables/tab_design_scope.tex
\begin{table}[t]
\centering
\caption{Agent-side reconstruction assumptions and scale treatment in representative
cooperative mapping systems.}
\label{tab:design_scope}
\vspace{-2.0mm}

\footnotesize
\setlength{\tabcolsep}{3pt}
\renewcommand{\arraystretch}{1.05}
\begin{tabular}{
>{\raggedright\arraybackslash}p{0.28\columnwidth}
>{\raggedright\arraybackslash}p{0.36\columnwidth}
>{\raggedright\arraybackslash}p{0.28\columnwidth}}
\toprule
System class &
Agent-side reconstruction assumption &
Scale treatment \\
\midrule
Metric cooperative SLAM~\cite{patel2023covins,tian2022kimeramulti,lajoie2024swarmslam} &
metric local estimates from visual--inertial, stereo, or range sensing &
shared metric scale \\
\midrule
Shared-pipeline monocular collaboration~\cite{zhou2025multi,cao2026como3r,gong2026magist3r} &
common learned reconstruction pipeline across agents &
monocular gauges coordinated within the common pipeline \\
\midrule
\textbf{MR.ScaleMaster} &
heterogeneous front-ends via session-local packets &
independent session scale conventions aligned during fusion \\
\bottomrule
\end{tabular}
\vspace{-5mm}
\end{table}

%% file: src_v3/system_overview.tex
\section{System Overview}
\label{sec:system}

MR.ScaleMaster follows a centralized architecture
(\figref{fig:system_overview}). Each monocular front-end produces
session-local keyframes, while the server applies \acs{SPC} to initialize
session scales when
heterogeneous scale groups are detected, constructs \acs{CFL}s, and
performs global $\mathrm{Sim}(3)$ anchor-node pose-graph optimization. For
session $r$, keyframe $i$ is provided to the server as
\begin{equation}
    \mathcal{K}_i^r =
    \bigl(I_i^r,\mathbf{X}_i^r,\mathbf{P}_i^r\bigr),
    \label{eq:packet}
\end{equation}
where $I_i^r$ is the image, $\mathbf{X}_i^r\in\mathrm{Sim}(3)$ is the
session-local pose, and $\mathbf{P}_i^r$ is its per-pixel point map. We assume
that $\mathbf{P}_i^r$ contains camera-frame 3D points; outputs in another
frame are converted before packetization. We further assume
that $\mathbf{X}_i^r$ and $\mathbf{P}_i^r$ share the same unknown session scale
$\lambda_r$, such that their translational and point-map coordinates are scaled
by $\lambda_r$ relative to a common reference geometry
(\tabref{tab:packet}).

A server-side matcher (MASt3R) supplies cross-session correspondences, from
which \secref{sec:loop_factor} builds \acs{CFL}s for the anchor-node graph of
\secref{sec:sim3_anchor}.

\input{src_v3/tables/tab_packet}

%% file: src_v3/tables/tab_packet.tex
\begin{table}[t]
\centering
\caption{Keyframe packet provided by session $r$.}
\label{tab:packet}
\vspace{-2.0mm}

\small
\setlength{\tabcolsep}{4pt}
\renewcommand{\arraystretch}{1.0}
\begin{tabular}{@{}ccl@{}}
\toprule
Symbol & Scale & Quantity \\
\midrule
$I_i^r$          & --          & keyframe image \\
$\mathbf{X}_i^r$ & $\lambda_r$ & session-local pose \\
$\mathbf{P}_i^r$ & $\lambda_r$ & per-pixel point map \\
\bottomrule
\end{tabular}

\vspace{-5mm}
\end{table}

%% file: src_v3/method.tex
\section{Method}
\label{sec:method}
\input{src_v3/figures/fig_multi_floor}
\input{src_v3/method_self_calibration}
\input{src_v3/method_retrieval}
\input{src_v3/method_loop_factor}
\input{src_v3/figures/fig_ablation_se3_sim3}
\subsection{\texorpdfstring{$\mathrm{Sim}(3)$}{Sim(3)} Anchor-Node Pose-Graph Optimization}
\label{sec:sim3_anchor}
\input{src_v3/method_multi_edge_jacob_formulation}
\noindent \textbf{Anchor-only vs.\ full $\mathbf{Sim(3)}$ optimization:} In
\emph{anchor-only} mode, keyframes are fixed and only $\{\mathbf{S}_r\}$ are
optimized; this solution initializes subsequent full $\mathrm{Sim}(3)$
refinement over anchors and keyframes, as illustrated in
\figref{fig:se3_sim3}.
\input{src_v3/method_scale_alarm}

%% file: src_v3/figures/fig_multi_floor.tex
\begin{figure}[!b]
    \centering
    \includegraphics[width=1.0\columnwidth]{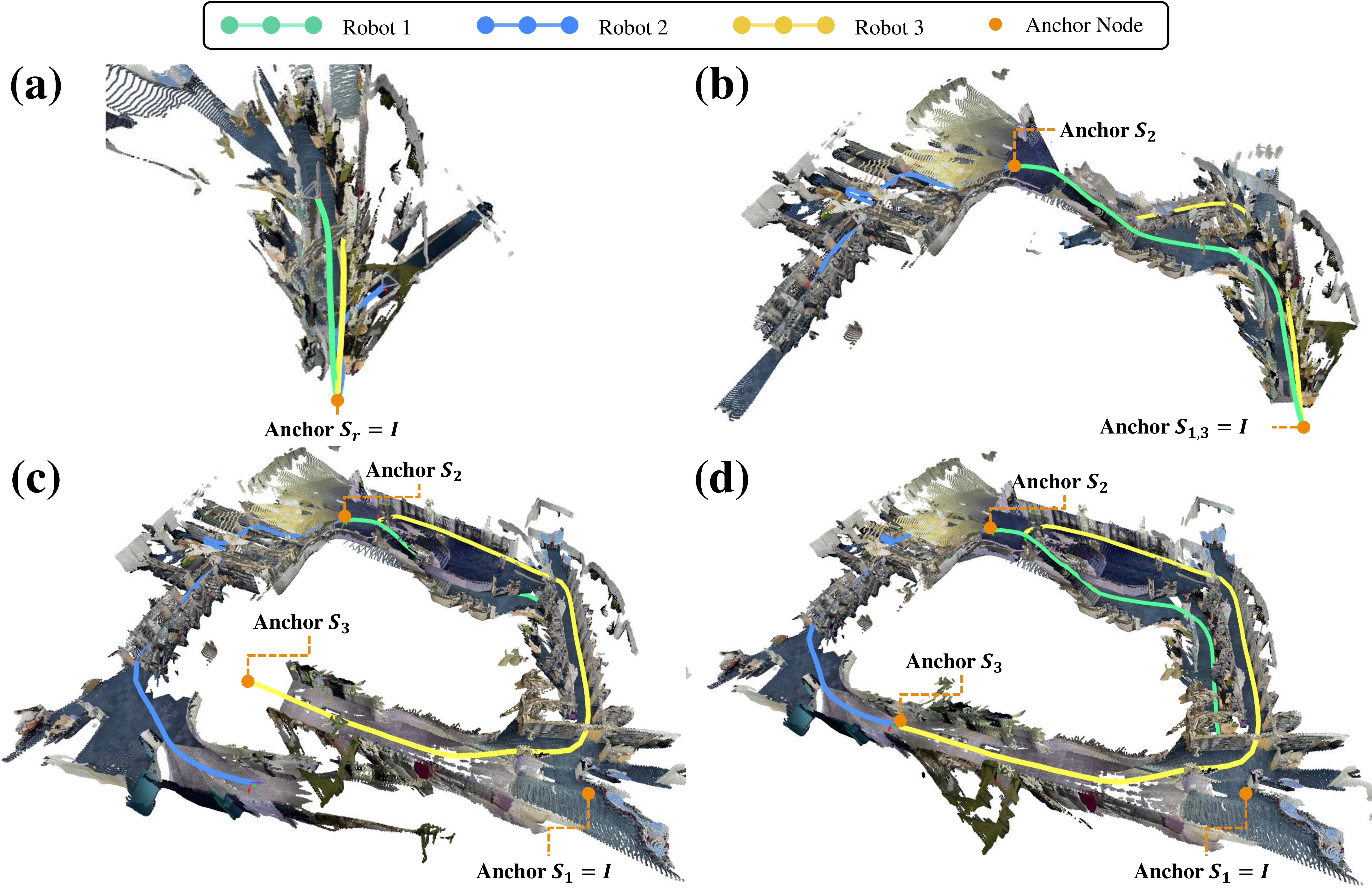}
\caption{Incremental $\mathrm{Sim}(3)$ anchor alignment for three sessions (green, blue, and yellow; orange: anchors; $\mathbf{S}_1=\mathbf{I}$). (a) Unaligned local maps. (b--c) Inter-session loop closures progressively align the sessions. (d) Full $\mathrm{Sim}(3)$ optimization yields the fused map.}   
    \label{fig:toy_sim3_anchor}
\end{figure}

%% file: src_v3/method_self_calibration.tex
\subsection{Scale Preconditioning (SPC)}
\label{sec:self_calibration}
\vspace{-1.5mm}
\acs{SPC} initializes per-session scales for anchor-graph
optimization. For each
of the first $M=5$ keyframes, we fit $\rho_{r,i}$ under
$d_{\mathrm{DA3}}\simeq\rho_{r,i}d_{\mathrm{FE}}$ using ray-matched
camera-axis depths from the front-end point map and Depth-Anything-3
prediction~\cite{da3}, and set $\rho_r=\operatorname{median}_i\rho_{r,i}$.
Across all seven integrated front-ends, these estimates differ by up to
$123\times$, so a unit-scale anchor
initialization may lie far from the solution when cross-session constraints
are sparse; the anchor $\mathbf{S}_r$ absorbs the remaining error.

Let $r_0$ be the fixed reference session. To avoid perturbing homogeneous
fleets, we sort $\{\rho_r\}$ and split at multiplicative gaps above $\tau_c$,
yielding clusters
$\{\mathcal{G}_l\}_{l=1}^{L}$. The correction is
\begin{equation}
    c_r =
    \begin{cases}
        1, & L = 1 \\[2pt]
        \operatorname{median}_{q \in \mathcal{G}(r)} \rho_q \;\big/\;
        \operatorname{median}_{q \in \mathcal{G}(r_0)} \rho_q, & L > 1 ,
    \end{cases}
    \label{eq:precal_cluster}
\end{equation}
where $\tau_c=1.6$ lies between the largest within-fleet gap ($1.55\times$)
and the smallest inter-front-end gap ($1.75\times$) observed across the five
heterogeneous KITTI fleets; any $\tau_c$ in this interval yields the same
clustering. Before factor construction, we multiply pose translations and
point-map coordinates by $c_r$, leaving rotations unchanged. Henceforth,
$\lambda_r$ is the residual session scale after this correction.

%% file: src_v3/method_retrieval.tex
\subsection{Loop Retrieval}
\label{sec:retrieval}

The server queries a shared image index using MASt3R's retrieval
head~\mbox{\cite{leroy2024mast3r}} to propose cross-session candidates.
Because retrieval uses images rather than reconstructed geometry, it is
unaffected by session scale conventions;
candidates are accepted only after correspondence-density verification.

%% file: src_v3/method_loop_factor.tex
\subsection{Cross-Front-End Loop Factor (CFL)}
\label{sec:loop_factor}

\noindent \textbf{Motivation:} MASt3R-SLAM estimates a relative $\mathrm{Sim}(3)$ from MASt3R pairwise point maps using the ray residual in its Eq.~(6), augmented by a low-weight residual on distance from the camera center and solved with the Gauss--Newton update in its Eq.~(7)~\cite{murai2025mast3r}. We refer to this released construction as RayOpt. RayOpt is appropriate when its pairwise reconstruction and the session-local graph geometry share a scale convention.

Across heterogeneous front-ends, however, the contributing sessions retain residual scales $\lambda_r$ and $\lambda_q$ after \acs{SPC}, whereas MASt3R produces an independent pairwise reconstruction in scale $\lambda_{\mathrm{m}}$. The resulting factor therefore has translation in $\lambda_{\mathrm{m}}$ units and a near-unity scale, rather than the cross-session ratio $\lambda_q/\lambda_r$. An anchor can absorb a uniform scale discrepancy, but MASt3R's pairwise maps are independent reconstructions rather than rescaled copies of the packet point maps, so their disagreement varies across points and also biases translation.

\noindent \textbf{Packet-geometry correspondences:} MASt3R supplies pixel correspondences $(\mathbf{u}_n, \mathbf{v}_n)$ and confidences $\alpha_n$ for a candidate pair $(I^r_i, I^q_j)$. Because preprocessing changes image dimensions, the packet point-map grids need not match the MASt3R grid. Here, $\pi_r$ and $\pi_q$ deterministically invert the corresponding center-crop-and-resize mappings using only the respective grid dimensions; they require no front-end-specific calibration or metadata. We form the 3D--3D correspondences
\begin{equation}
    \mathbf{p}_n = \mathbf{P}^r_i\!\left[\pi_r(\mathbf{u}_n)\right],
    \qquad
    \mathbf{q}_n = \mathbf{P}^q_j\!\left[\pi_q(\mathbf{v}_n)\right],
    \label{eq:lift_own_depth}
\end{equation}
which are expressed in the residual session scales $\lambda_r$ and
$\lambda_q$, respectively. We discard matches for which either indexed 3D
point is invalid or the correspondence confidence falls below a threshold.

\noindent \textbf{Robust $\mathbf{Sim(3)}$ estimation:} Let $\mathbf{Z}_{j\leftarrow i}$ map $\mathbf{p}_n$ to $\mathbf{q}_n$, and define $\mathbf{p}'_n=s\mathbf{R}\mathbf{p}_n+\mathbf{t}$. Gross outliers are removed by RANSAC with a scale-enabled point-to-point estimator, and the inlier set is refined by a confidence-weighted ray-and-distance residual that separates bearing information from noisier range estimates. With $\hat{\mathbf{x}} = \mathbf{x} / \|\mathbf{x}\|$,
\begin{align}
    \widetilde{\mathbf{Z}}_{j\leftarrow i}
    &= \arg\min_{\mathbf{Z} \in \mathrm{Sim}(3)} \sum_{n} \alpha_n
    \Bigg[\frac{\left\| \hat{\mathbf{q}}_n - \hat{\mathbf{p}}'_n \right\|^2}{\sigma_{\mathrm{ray}}^2} \nonumber \\
    &\hspace{27mm}+
    \frac{\left( \|\mathbf{q}_n\| - \|\mathbf{p}'_n\| \right)^2}{\sigma_{\mathrm{dist}}^2}\Bigg],
    \label{eq:loop_sim3_objective}
\end{align}
solved by Gauss--Newton on $\mathfrak{sim}(3)$; $\sigma_{\mathrm{ray}}$ and $\sigma_{\mathrm{dist}}$ weight this estimator only, not the pose graph. In \eqref{eq:loop_sim3_objective}, the ray term is invariant to a common rescaling of $s$ and $\mathbf{t}$, and therefore constrains $\mathbf{R}$ and $\mathbf{t}/s$, but not the absolute scale. The bearing component supplies scale-invariant rotational information, so we retain $\tilde{\mathbf{R}}$ from this stage. Holding $\mathbf{R}=\tilde{\mathbf{R}}$ fixed, we then re-estimate scale and translation from the 3D residual using the weighted Umeyama solution~\cite{umeyama1991}:
\begin{equation}
    s^{\star} = \frac{\sum_n w_n \left\langle \mathbf{q}_n - \bar{\mathbf{q}},\; \tilde{\mathbf{R}}\mathbf{p}_n - \bar{\mathbf{a}} \right\rangle}
                      {\sum_n w_n \left\| \tilde{\mathbf{R}}\mathbf{p}_n - \bar{\mathbf{a}} \right\|^2},
    \qquad
    \mathbf{t}^{\star} = \bar{\mathbf{q}} - s^{\star}\,\bar{\mathbf{a}},
    \label{eq:scale_translation}
\end{equation}
where $\bar{\mathbf{a}}$ and $\bar{\mathbf{q}}$ are the weighted centroids of $\{\tilde{\mathbf{R}}\mathbf{p}_n\}$ and $\{\mathbf{q}_n\}$, and $w_n$ are Huber weights seeded by $\alpha_n$ and re-evaluated on the residual $\mathbf{q}_n - s\tilde{\mathbf{R}}\mathbf{p}_n - \mathbf{t}$. A candidate with fewer than three inliers or an ill-conditioned system is rejected. For an accepted candidate, the final loop measurement is $\mathbf{Z}_{j\leftarrow i} = (\tilde{\mathbf{R}}, \mathbf{t}^{\star}, s^{\star})$. Because $\mathbf{p}_n$ and $\mathbf{q}_n$ are expressed in the residual session scales $\lambda_r$ and $\lambda_q$, respectively, we have $s^{\star} \approx \lambda_q/\lambda_r$ with $\mathbf{t}^{\star}$ in the target session's units. Thus, $\mathbf{Z}_{j\leftarrow i}$ requires no additional rescaling before insertion into the graph.

%% file: src_v3/figures/fig_ablation_se3_sim3.tex
\begin{figure}[t]
    \centering
    \includegraphics[width=1.0\columnwidth]{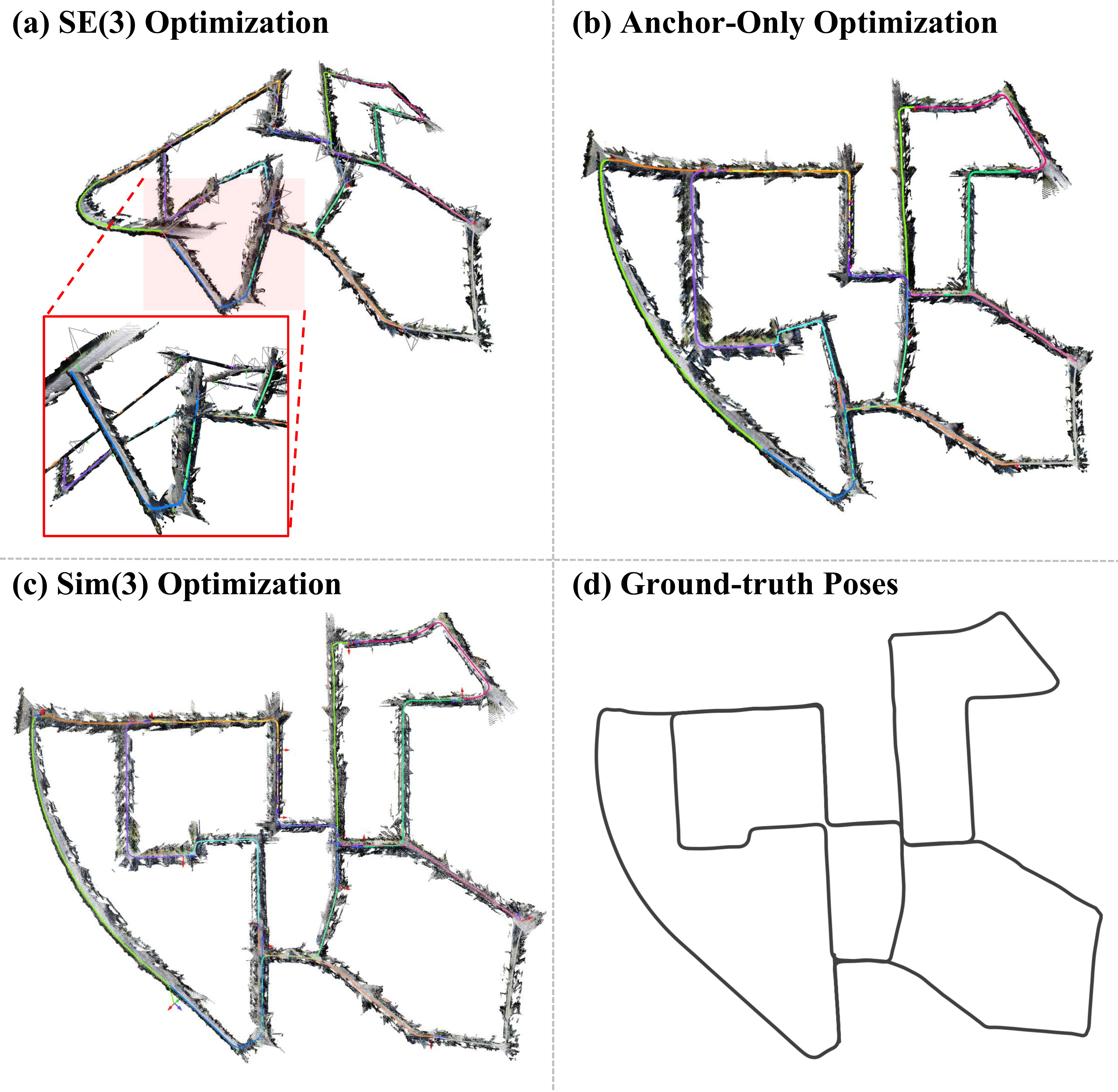}
\caption{Scale-aware backend ablation on \texttt{KITTI 00} with
15 sessions. (a) $\mathrm{SE}(3)$ optimization absorbs inter-session scale
inconsistency as boundary distortions. (b) Anchor-only $\mathrm{Sim}(3)$
aligns sessions while preserving local trajectories. (c) Full
$\mathrm{Sim}(3)$ optimization jointly refines anchors and keyframes.
(d) Ground truth. Quantitative results are reported in \tabref{tab:ablation}.}
\label{fig:se3_sim3}
\vspace{-5mm}
\end{figure}

%% file: src_v3/method_multi_edge_jacob_formulation.tex
\noindent \textbf{Multi-robot pose-graph formulation:} For each session $r$, keyframe poses $\mathbf{X}^r_i \in \mathrm{Sim}(3)$ are expressed in the session-local frame, and an anchor $\mathbf{S}_r \in \mathrm{Sim}(3)$ maps that frame into the global frame, as illustrated in \figref{fig:toy_sim3_anchor}:
\begin{equation}
    \mathbf{T}^r_i = \mathbf{S}_r \cdot \mathbf{X}^r_i
    \label{eq:world_pose}
\end{equation}
Intra-session odometry edges constrain consecutive keyframes, while inter-session loop edges carry the directed measurement $\mathbf{Z}_{j\leftarrow i}$ constructed in \secref{sec:loop_factor}. We fix $\mathbf{S}_1 = \mathbf{I}$ to remove the global gauge.

\noindent \textbf{Inter-session loop closure error:} The predicted $i\!\to\!j$ transformation is $(\mathbf{T}^q_j)^{-1}\mathbf{T}^r_i$; thus,
\begin{equation}
    \mathbf{e} = \mathrm{Log}\!\left(\mathbf{Z}_{j\leftarrow i}^{-1} \cdot (\mathbf{S}_q \cdot \mathbf{X}^q_j)^{-1} \cdot \mathbf{S}_r \cdot \mathbf{X}^r_i\right) .
    \label{eq:error_expanded}
\end{equation}
Each inter-session factor is weighted by
$\boldsymbol{\Omega}=\operatorname{diag}(\sigma_R^{-2}\mathbf{I}_3,
\sigma_t^{-2}\mathbf{I}_3,\sigma_s^{-2})$, where $\sigma_R$, $\sigma_t$,
and $\sigma_s$ are tuned standard-deviation parameters for the rotation,
translation, and log-scale residual blocks, respectively.

\noindent \textbf{Analytic Jacobians:}
Using the $\mathrm{Sim}(3)$ group adjoint and the right Jacobian inverse on $\mathfrak{sim}(3)$~\cite{strasdat2012local}, we obtain the four Jacobian blocks of~\eqref{eq:error_expanded} analytically. The group adjoint $\mathrm{Ad}_{\mathbf{T}}$ of a $\mathrm{Sim}(3)$ element $\mathbf{T} = (\mathbf{R}, \mathbf{t}, s)$ is
\begin{equation}
    \mathrm{Ad}_{\mathbf{T}} =
    \begin{bmatrix}
        \mathbf{R} & \mathbf{0} & \mathbf{0} \\
        [\mathbf{t}]_\times \mathbf{R} & s\mathbf{R} & -\mathbf{t} \\
        \mathbf{0}^\top & \mathbf{0}^\top & 1
    \end{bmatrix}
    \label{eq:adjoint}
\end{equation}
where $[\mathbf{t}]_\times$ denotes the skew-symmetric matrix of $\mathbf{t}$. The Lie algebra adjoint $\mathrm{ad}_{\boldsymbol{\xi}}$ for $\boldsymbol{\xi} = (\boldsymbol{\omega}, \boldsymbol{\upsilon}, \sigma)$ is:
\begin{equation}
    \mathrm{ad}_{\boldsymbol{\xi}} =
    \begin{bmatrix}
        [\boldsymbol{\omega}]_\times & \mathbf{0} & \mathbf{0} \\
        [\boldsymbol{\upsilon}]_\times & [\boldsymbol{\omega}]_\times + \sigma \mathbf{I} & -\boldsymbol{\upsilon} \\
        \mathbf{0}^\top & \mathbf{0}^\top & 0
    \end{bmatrix}
    \label{eq:lie_adjoint}
\end{equation}
The right Jacobian inverse $\mathbf{J}_r^{-1}(\boldsymbol{\xi})$ is evaluated by its Bernoulli series, truncated at sixth order:
\begin{equation}
    \mathbf{J}_r^{-1}(\boldsymbol{\xi}) = \mathbf{I} + \tfrac{1}{2}\,\mathrm{ad}_{\boldsymbol{\xi}} + \tfrac{1}{12}\,\mathrm{ad}_{\boldsymbol{\xi}}^2 - \tfrac{1}{720}\,\mathrm{ad}_{\boldsymbol{\xi}}^4 + \tfrac{1}{30240}\,\mathrm{ad}_{\boldsymbol{\xi}}^6 .
    \label{eq:jr_inv}
\end{equation}
With $\mathbf{e}$ as in~\eqref{eq:error_expanded}, the Jacobians with respect to left perturbations on the source pose $\mathbf{T}^r_i$ and target pose $\mathbf{T}^q_j$ are:
\begin{equation}
    \mathbf{J}_{\mathbf{T}^r_i} = \mathbf{J}_r^{-1}(\mathbf{e})\,\mathrm{Ad}_{(\mathbf{T}^r_i)^{-1}}, \quad
    \mathbf{J}_{\mathbf{T}^q_j} = -\mathbf{J}_r^{-1}(\mathbf{e})\,\mathrm{Ad}_{(\mathbf{T}^r_i)^{-1}}
    \label{eq:jac_world}
\end{equation}
Applying the chain rule for the composite $\mathbf{T}^r_i = \mathbf{S}_r \cdot \mathbf{X}^r_i$, a left perturbation on $\mathbf{S}_r$ directly perturbs $\mathbf{T}^r_i$, while a left perturbation on $\mathbf{X}^r_i$ is mapped through $\mathrm{Ad}_{\mathbf{S}_r}$. The four Jacobian blocks for the 4-vertex edge are:
\begin{align}
    \mathbf{J}_{\mathbf{S}_r} &= \mathbf{J}_{\mathbf{T}^r_i}, \quad
    \mathbf{J}_{\mathbf{X}^r_i} = \mathbf{J}_{\mathbf{T}^r_i} \, \mathrm{Ad}_{\mathbf{S}_r}, \nonumber \\
    \mathbf{J}_{\mathbf{S}_q} &= \mathbf{J}_{\mathbf{T}^q_j}, \quad
    \mathbf{J}_{\mathbf{X}^q_j} = \mathbf{J}_{\mathbf{T}^q_j} \, \mathrm{Ad}_{\mathbf{S}_q}
    \label{eq:jac_vertices}
\end{align}

%% file: src_v3/method_scale_alarm.tex
\subsection{Scale Collapse Alarm (SCA)}
\label{sec:scale_alarm}

A spurious intra-session loop can be absorbed by the scale degree of freedom in a
$\mathrm{Sim}(3)$ optimizer, causing abrupt scale collapse or explosion that
the cooperative backend cannot reliably repair. We therefore validate loop
candidates locally using two gates.

\noindent\textbf{Turning gate:}
A candidate spanning more than a minimum keyframe gap is rejected when the
intervening trajectory exhibits insufficient net turning. Signed per-step
yaw is accumulated about an online-estimated turning axis: back-and-forth
motion cancels, whereas a genuine circuit retains its accumulated turning.
This gate is evaluated before factor insertion.

\noindent\textbf{Scale-change gate:}
Each remaining candidate is inserted tentatively and rolled back if its
induced per-keyframe scale change exceeds $\mu+3\sigma$, where $\mu$ and
$\sigma$ are running statistics of the scale change on keyframe pairs
unaffected by loop candidates. The threshold is mildly relaxed for
candidates with large accumulated rotation, and gross per-keyframe scale
jumps trigger a rollback regardless; all thresholds are in the released
configuration.

\acs{SCA} is applicable to front-ends that expose a per-keyframe scale
trajectory and support factor rollback; our implementation uses the
MASt3R-SLAM incremental optimizer. The server-side backend remains
independent of the alarm.

%% file: src_v3/experiments.tex
\section{Experiments}
\label{sec:exp}

\input{src_v3/figures/fig_scale_alarm}
\input{src_v3/tables/tab_ablation}
\input{src_v3/figures/fig_kitti_020507}
\input{src_v3/tables/tab_mixed_correction}
\input{src_v3/tables/tab_loop_measurement_ablation}
\input{src_v3/tables/tab_kitti_ate}
\input{src_v3/tables/tab_frontend_agnostic}
\input{src_v3/tables/tab_scale_alarm_full_stack}

\subsection{Experimental Setup}
\label{sec:exp_setup}

\noindent\textbf{Datasets and session generation:} We evaluate on KITTI
odometry sequences 00, 02, 05, 07, and 08~\cite{geiger2013vision}, spanning
0.7--5.1\,km, and the UT Campus Object Dataset (CODa)~\cite{zhang2024toward}.
For KITTI, each sequence is partitioned into $N\in\{1,5,10,15\}$
non-overlapping sub-sequences, treated as independent sessions.
 CODa sequences (up to 20,000 images) are sparsified into ORB-selected keyframes shared by all methods. Single-session runs process them as one sequence, whereas
MR.ScaleMaster partitions them into five partially overlapping sessions
($\sim$10\% overlap). Partitioning a single recording emulates crowd-sourced
collection under controlled conditions; \figref{fig:motivation} and
\figref{fig:coda_multi_robot_bev} use independently recorded CODa sequences
from different days.

\noindent\textbf{Front-ends and metrics:} The session-count experiment
(\tabref{tab:scalability}) uses MASt3R-SLAM~\cite{murai2025mast3r}, and
the heterogeneous KITTI experiment (\tabref{tab:mixed_correction}) combines
MASt3R-SLAM, $\pi^3$~\cite{wang2025pi3}, and
VGGT-SLAM~2.0~\cite{maggio2025vggt-slam2}.
The loop-measurement comparison (\tabref{tab:loop_measurement_ablation}) adds
CUT3R, TTT3R, LoGeR, and LingBot-Map; the CODa experiments
(\secref{sec:exp_frontend}) use these four and VGGT-SLAM~2.0.
Each session provides $\mathcal{K}^r_i$ from~\eqref{eq:packet}. We report
\ac{ATE} RMSE using \texttt{evo}~\cite{grupp2017evo}, fitting one global
$\mathrm{Sim}(3)$ Umeyama alignment jointly to all sessions in each fused
result; DIV denotes divergence.
Because $\mathrm{Sim}(3)$-aligned ATE does not directly expose inter-session
scale consistency, we additionally define from raw path lengths the
GT-derived relative scale that maps session $r$ into the units of the
reference session $r_0$,
\begin{equation}
    s^{\mathrm{GT}}_{r\to r_0} =
    \frac{\ell^{\mathrm{GT}}_r / \ell_r}{\ell^{\mathrm{GT}}_{r_0} / \ell_{r_0}},
    \label{eq:gt_scale}
\end{equation}
where $\ell_r$ and $\ell^{\mathrm{GT}}_r$ are the raw estimated and
ground-truth path lengths evaluated at identical keyframe timestamps. Let
$\hat{\gamma}_r$ denote the scale component of the optimized anchor
$\hat{\mathbf{S}}_r$ after full graph optimization. With preconditioning
factor $c_r$, the estimated
relative scale is $\hat{s}_{r\to r_0}=c_r\hat{\gamma}_r$ under the reference
gauge $c_{r_0}=\hat{\gamma}_{r_0}=1$, where $r_0=1$ is the fixed anchor. We
report
\begin{equation}
    E_s = \frac{1}{R-1}\sum_{r\neq r_0}
    \left|\log\frac{\hat{s}_{r\to r_0}}{s^{\mathrm{GT}}_{r\to r_0}}\right|,
    \label{eq:scale_error}
\end{equation}
where $R$ is the number of sessions and lower values indicate better
inter-session scale consistency.

\noindent\textbf{Implementation and parameters:} We optimize the graph in
g2o~\cite{kummerle2011g} using the proposed multi-vertex $\mathrm{Sim}(3)$
edge. All experiments run on a
desktop with an AMD Ryzen~9 9900X CPU, 64\,GB RAM, and an NVIDIA RTX 5090 GPU.
The inter-session standard-deviation parameters $(\sigma_R,\sigma_t,\sigma_s)$
are selected once by TPE~\cite{akiba2019optuna} on MASt3R-SLAM over KITTI to
minimize mean ATE, then fixed for all other front-ends and CODa. The resulting
configuration will be released with the code.

\input{src_v3/figures/fig_coda_multi_robot_bev}

\noindent\textbf{Runtime and data footprint:} The uncompressed $512\times384$
float32 point map occupies 2.25\,MiB per keyframe, or 3.2--11.7\,MiB/s at
1.4--5.2 keyframes/s on KITTI. Packets are replayed from disk, so this is an
interface estimate, not measured network traffic. On one RTX 5090, ingestion
and retrieval take 57\,ms per keyframe (median; 34\,ms MASt3R, 17\,ms
retrieval), excluding candidate-dependent processing. Per-candidate matching
and factor estimation take 23\,ms and 90--140\,ms; full optimization takes
0.27\,s (max 0.47\,s) for 15 sessions and 1357 keyframes.

\noindent\textbf{Baselines:} The geometry-source ablations use RayOpt, the
MASt3R-SLAM ray-and-distance construction defined in
\secref{sec:loop_factor}, as a reference rather than a complete heterogeneous
fusion baseline. Of the closest shared-pipeline systems,
Multi-Agent Monocular Dense SLAM~\cite{zhou2025multi} has no public code, and
CoMo3R-SLAM~\cite{cao2026como3r} did not recover from repeated relocalization
on our sparse CODa keyframes. On its native two-agent Barn example (55\,m,
20 overlapping frames), CoMo3R-SLAM obtains 0.084\,m ATE versus 0.15, 0.12,
and 0.13\,m for MR.ScaleMaster with MASt3R-SLAM, LoGeR, and LingBot-Map under
the same $\mathrm{Sim}(3)$ alignment. This shared-pipeline setting favors
CoMo3R-SLAM, whereas ours accepts each front-end unchanged.
\subsection{CFL and Scale-Aware Fusion}
\label{sec:exp_ablation}

\noindent\textbf{Scale-consistent fusion:} \tabref{tab:mixed_correction}
evaluates the contributions of \acs{CFL} and \acs{SPC} under GT-derived
path-length scale ratios of $57$--$103\times$.
Moving from A to B replaces RayOpt with \acs{CFL}, which reads its 3D points
from the input packets so that the factor measures the inter-session ratio
instead of a near-unity scale in matcher units: mean $E_s$ drops from
1.083 to 0.429 and mean ATE from 151.23 to 95.37\,m. Moving from B to C adds
\acs{SPC}, which provides a better initialization when a unit-scale anchor
initialization lies far from the solution
(\secref{sec:self_calibration}): $E_s$ decreases to 0.246 and ATE to
39.82\,m. The largest gains occur on \texttt{KITTI~02} and \texttt{08}, where
(B) leaves the largest scale error. \texttt{KITTI~07}, where (B) already attains
the lowest $E_s$, is the only exception ($E_s$: 0.136 to 0.218; ATE: 6.39 to
10.85\,m): with little scale error left to correct, error in the depth-prior
estimate can dominate.

\noindent\textbf{CFL across front-ends:}
\tabref{tab:loop_measurement_ablation} tests \acs{CFL} on 15-session,
same-front-end runs. Relative to RayOpt, with all other
components fixed, using each input packet's own 3D geometry improves 27 of 30
runs across the six
front-ends whose reconstruction models differ from MASt3R.
For MASt3R-SLAM, three of five runs improve, while the mean change is
$+3.64$\,m, dominated by \texttt{KITTI~08}.

\noindent\textbf{Scale-aware backend:}
On \texttt{KITTI~00} partitioned into 15 sessions, full $\mathrm{Sim}(3)$ optimization
reduces ATE from 137.26 to 16.61\,m relative to the $\mathrm{SE}(3)$ backend
($8.3\times$); anchor-only optimization reaches 18.92\,m and initializes full
refinement (\figref{fig:se3_sim3}, \tabref{tab:ablation}).
\figref{fig:kitti_qualitative} provides qualitative results across four
KITTI sequences.
\subsection{Effect of Session Count}
\label{sec:exp_scalability}

\tabref{tab:scalability} evaluates ATE versus session count for MASt3R-SLAM.
The default single-session configuration loses tracking on every sequence;
relaxing its 3D-correspondence threshold restores tracking but leaves drift.
Against this tuned baseline, the best multi-session settings reduce ATE by
$6.2\times$ and $13.6\times$ on \texttt{KITTI~00} and \texttt{05}.
\texttt{KITTI~02} is best with 10 sessions, while \texttt{KITTI~08} remains
limited by reverse-viewpoint retrieval. \texttt{KITTI~07} is worse at every
multi-session count: its short trajectory returns to its starting area, and
the tuned single-session run already closes its drift.

\subsection{Heterogeneous Front-End Deployment}
\label{sec:exp_frontend}

With parameters fixed after MASt3R-SLAM/KITTI tuning,
\tabref{tab:frontend_agnostic_ate} evaluates five front-ends on identical CODa
keyframes. Five-session fusion lowers mean ATE for every front-end and improves
32 of 35 pairs, although this gain combines shorter sessions with
inter-session fusion. Single-session long-context systems obtain mean ATE of 83--98\,m on
the same keyframes.
\figref{fig:coda_multi_robot_bev} fuses three CODa sequences reconstructed by
three different front-ends. Relative to separate five-session fusion, joint
optimization of all 15 sessions lowers ATE on all three sequences, from a
mean of 11.44 to 9.19\,m, and reduces the spread of the per-sequence
$\mathrm{Sim}(3)$ alignment scales from $11\times$ to $1.2\times$, placing
the three front-ends in one gauge.
\figref{fig:hero_shot} qualitatively demonstrates a multi-floor map from four
agents: a quadruped, a wheeled robot, and two handheld cameras; no ground truth
is available.

\subsection{Evaluation of SCA}
\label{sec:exp_alarm}

\tabref{tab:exp_alarm} evaluates \acs{SCA} on 22 ScaleMaster
segments~\cite{ju2026have}. \emph{Straight} and \emph{Rollback} test the turning and
scale-change gates, respectively; \emph{TP} contains genuine revisits. All
runs use MASt3R-SLAM and a permissive retrieval threshold that produces
batches containing both false-positive and genuine loops.
\figref{fig:scale_alarm} shows a representative segment.

\acs{SCA} recovers one divergent run to 8.44\,m and improves 14 segments by up
to 22.64\,m; one is unchanged and six degrade by at most 2.31\,m. In a mixed
batch, a false-positive loop can dominate the induced scale change and reject
co-occurring genuine loops, explaining these bounded degradations. The alarm
thus serves as a safeguard against scale collapse rather than a general
accuracy improvement.

%% file: src_v3/figures/fig_scale_alarm.tex
\begin{figure}[t]
    \centering
    \includegraphics[width=1.0\columnwidth]{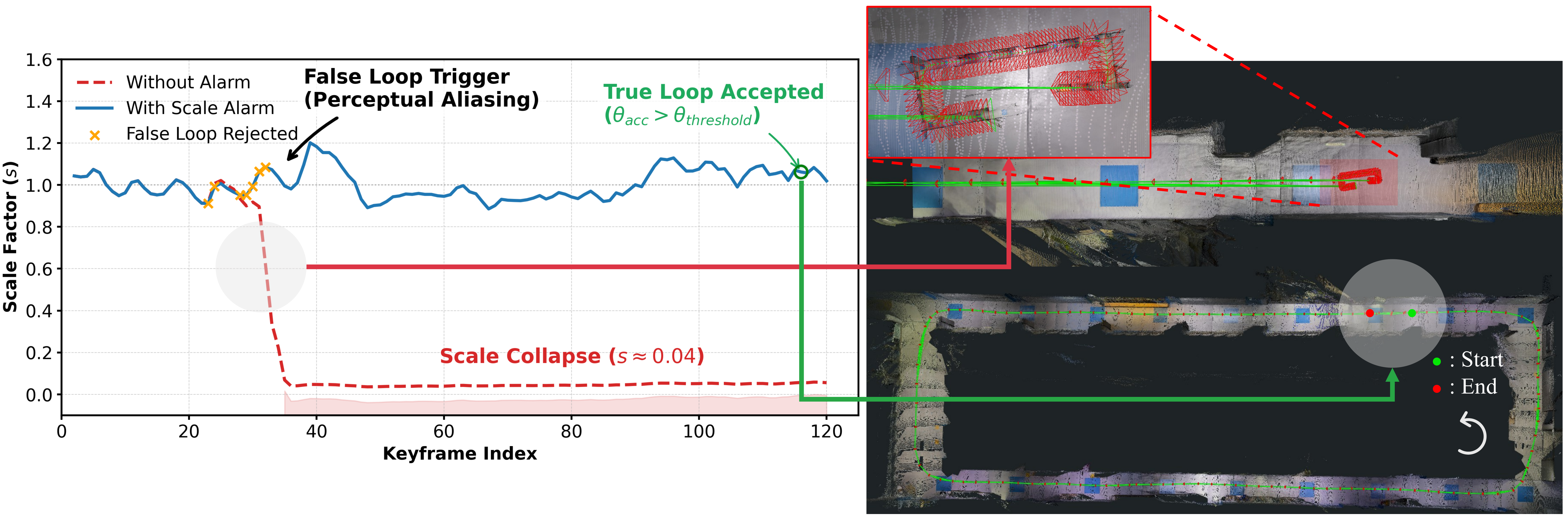}
    \caption{\acs{SCA} on a ScaleMaster segment with perceptual
    aliasing. Without the alarm, an aliasing-induced false loop collapses the
    MASt3R-SLAM session scale ($s\approx0.04$); with the alarm, the false loop
    is rejected and the genuine loop is accepted. Bottom: resulting maps
    without and with the alarm.}
    \label{fig:scale_alarm}
    \vspace{-3mm}
\end{figure}

%% file: src_v3/tables/tab_ablation.tex
\begin{table}[t!]
    \centering
    \caption{MASt3R-SLAM backend ablation on \texttt{KITTI 00} with 15 sessions.}
    \label{tab:ablation}
    \renewcommand{\arraystretch}{0.97}
    \vspace{-2mm}

    \setlength{\tabcolsep}{6pt}
    \begin{tabular}{@{}lc@{}}
        \toprule
        Configuration & ATE RMSE [m] \\
        \midrule
        $\mathrm{SE}(3)$ backend opt.          & 137.26 \\
        $\mathrm{Sim}(3)$ anchor-only opt.     & 18.92 \\
        $\mathrm{Sim}(3)$ full opt.            & \textbf{16.61} \\
        \bottomrule
    \end{tabular}
    \vspace{-2mm}

\end{table}

%% file: src_v3/figures/fig_kitti_020507.tex
\begin{figure}[t]
  \centering
  \begin{tabular*}{\linewidth}{@{\extracolsep{\fill}}cccc}
    \texttt{KITTI 02} & \texttt{KITTI 05} & \texttt{KITTI 07} & \texttt{KITTI 00}
  \end{tabular*}\vspace{-1mm}
  \includegraphics[width=\linewidth]{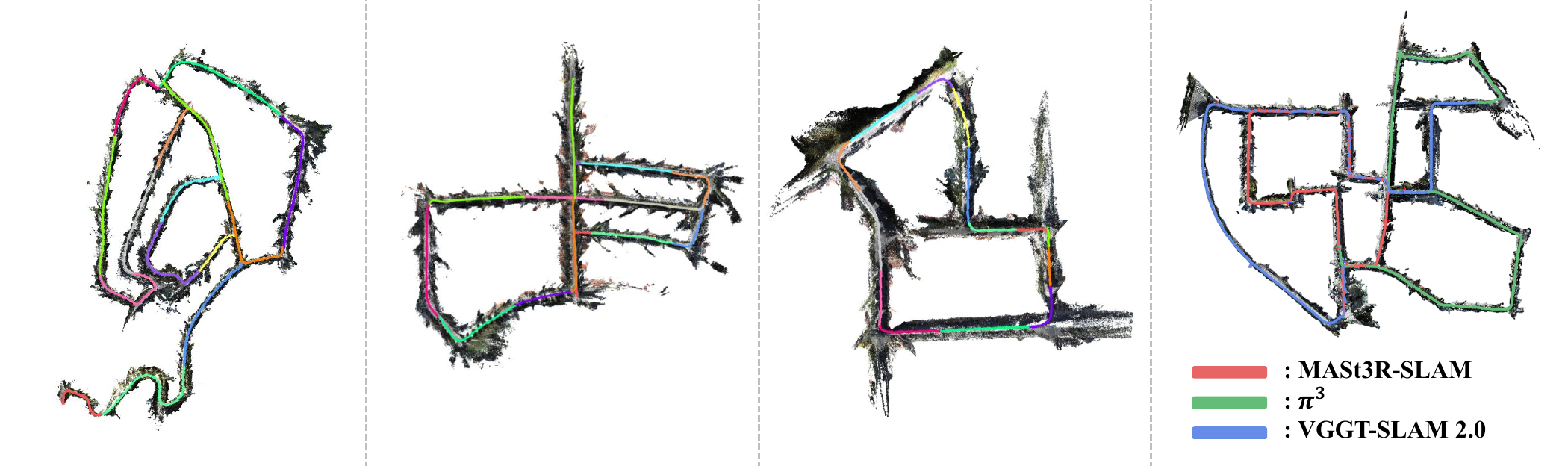}
  \caption{Qualitative multi-session fusion. LingBot-Map
  reconstructions on \texttt{KITTI 02}, \texttt{05}, and \texttt{07}
  (left to right) contain 15 aligned sessions. \texttt{KITTI 00} (right)
  combines five sessions each from MASt3R-SLAM, $\pi^3$, and VGGT-SLAM~2.0;
  colors denote front-end type.}
  \label{fig:kitti_qualitative}
  \vspace{-2mm}
\end{figure}

%% file: src_v3/tables/tab_mixed_correction.tex
\begin{table}[t!]
\centering
\caption{Heterogeneous KITTI ablation with five sessions each from
MASt3R-SLAM, $\pi^3$, and VGGT-SLAM~2.0; GT-derived path-length scale ratios
span $57$--$103\times$. ATE RMSE [m] and $E_s$ compare (A) RayOpt, the
MASt3R-SLAM ray-and-distance relative-pose construction described in its
Eqs.~(6)--(7)~\cite{murai2025mast3r}, (B) \acs{CFL}, and (C) \acs{CFL} with
\acs{SPC}. Lower is better.}
\label{tab:mixed_correction}
\vspace{-2.0mm}
\footnotesize
\setlength{\tabcolsep}{2.5pt}
\renewcommand{\arraystretch}{1.0}
\resizebox{\columnwidth}{!}{%
\begin{tabular}{@{}l ccc ccc@{}}
\toprule
 & \multicolumn{3}{c}{ATE RMSE [m] $\downarrow$}
 & \multicolumn{3}{c}{$E_s$ $\downarrow$} \\
\cmidrule(lr){2-4} \cmidrule(lr){5-7}
Seq & (A) RayOpt & (B) \acs{CFL} & \makecell{(C) \acs{CFL}\\$+$ \acs{SPC}}
& (A) & (B) & (C) \\
\midrule
\texttt{KITTI 00} & 134.78 & 25.05  & \textbf{19.43} & 1.285 & 0.153 & \textbf{0.133} \\
\texttt{KITTI 02} & 296.39 & 280.43 & \textbf{78.32} & 1.284 & 0.614 & \textbf{0.228} \\
\texttt{KITTI 05} & 56.22  & 18.52  & \textbf{9.79}  & 0.696 & 0.369 & \textbf{0.112} \\
\texttt{KITTI 07} & 21.20  & \textbf{6.39} & 10.85 & 0.808 & \textbf{0.136} & 0.218 \\
\texttt{KITTI 08} & 247.58 & 146.45 & \textbf{80.71} & 1.343 & 0.872 & \textbf{0.541} \\
\midrule
Mean              & 151.23 & 95.37 & \textbf{39.82} & 1.083 & 0.429 & \textbf{0.246} \\
\bottomrule
\end{tabular}%
}
\vspace{-5mm}
\end{table}

%% file: src_v3/tables/tab_loop_measurement_ablation.tex
\begin{table}[t]
\centering
\caption{\acs{CFL} ablation across seven front-ends on KITTI (15
same-front-end sessions per run). $\Delta$ATE [m] is \acs{CFL} minus RayOpt,
which builds factors from MASt3R pairwise point maps; all other components are
fixed. Negative values favor \acs{CFL}; bold marks
$\Delta\mathrm{ATE}\leq-10$\,m.}
\label{tab:loop_measurement_ablation}
\vspace{-2mm}
\scriptsize
\setlength{\tabcolsep}{2.0pt}
\renewcommand{\arraystretch}{1.0}
\begin{tabular}{@{}lrrrrrr@{}}
\toprule
Front-end & \texttt{00} & \texttt{02} & \texttt{05} & \texttt{07} & \texttt{08} & Mean \\
\midrule
MASt3R-SLAM & $-1.60$ & $-5.83$ & $+1.23$ & $-2.60$ & $+26.99$ & $+3.64$ \\
$\pi^3$ & $\mathbf{-19.46}$ & $\mathbf{-66.15}$ & $\mathbf{-12.71}$ & $\mathbf{-16.72}$ & $\mathbf{-19.91}$ & $\mathbf{-26.99}$ \\
CUT3R & $-4.95$ & $\mathbf{-76.10}$ & $-3.73$ & $-3.62$ & $\mathbf{-12.82}$ & $\mathbf{-20.24}$ \\
TTT3R & $-1.46$ & $\mathbf{-125.04}$ & $\mathbf{-10.27}$ & $-4.69$ & $\mathbf{-19.63}$ & $\mathbf{-32.22}$ \\
VGGT-SLAM~2.0 & $\mathbf{-10.73}$ & $+26.59$ & $+8.70$ & $-6.84$ & $\mathbf{-63.16}$ & $-9.09$ \\
LoGeR & $-4.66$ & $\mathbf{-37.67}$ & $+2.20$ & $-1.74$ & $\mathbf{-14.75}$ & $\mathbf{-11.32}$ \\
LingBot-Map & $\mathbf{-20.15}$ & $\mathbf{-184.24}$ & $\mathbf{-13.61}$ & $-6.76$ & $\mathbf{-28.29}$ & $\mathbf{-50.61}$ \\
\bottomrule
\end{tabular}%
\vspace{-2.0mm}
\end{table}

%% file: src_v3/tables/tab_kitti_ate.tex
\begin{table}[t!]
    \centering
    \caption{ATE RMSE [m] on KITTI~\cite{geiger2013vision} vs.\ session count
    using MASt3R-SLAM. The tuned single-session baseline relaxes the
    3D-correspondence acceptance threshold; all other settings are unchanged.
    Best per sequence in \textbf{bold}; TL denotes tracking loss.}
    \label{tab:scalability}
    \vspace{-2mm}
    \setlength{\tabcolsep}{3pt}
    \renewcommand{\arraystretch}{0.95}
    \resizebox{0.9\columnwidth}{!}{%
    \begin{tabular}{@{}lc|cc|ccc@{}}
        \toprule
        Sequence & Length & 1 & 1\,(tuned) & 5 & 10 & 15 \\
        \midrule
        \texttt{KITTI 00} & 3.7\,km & TL & 103.07 & 48.69  & 19.66  & \textbf{16.61} \\
        \texttt{KITTI 02} & 5.1\,km & TL & 158.49 & 113.36 & \textbf{83.06}  & 88.54 \\
        \texttt{KITTI 05} & 2.2\,km & TL & 147.15 & 13.33  & \textbf{10.82}  & 12.01 \\
        \texttt{KITTI 07} & 0.7\,km & TL & \textbf{2.04} & 7.05 & 9.79 & 8.67 \\
        \texttt{KITTI 08} & 3.2\,km & TL & 88.16  & 166.68 & 104.83 & \textbf{85.91} \\
        \midrule
        Mean              &         & -- & 99.78  & 69.82  & 45.63  & \textbf{42.35} \\
        \bottomrule
    \end{tabular}}
    \vspace{-5mm}
\end{table}

%% file: src_v3/tables/tab_frontend_agnostic.tex
\begin{table*}[t!]
\centering
\caption{CODa ATE RMSE [m] on identical ORB-selected keyframes. Paired cells
report five-session MR.ScaleMaster / same-front-end single-session results;
bold marks the lower value. The last three columns are single-session
long-context references.}
\label{tab:frontend_agnostic_ate}
\vspace{-2.0mm}
\footnotesize
\setlength{\tabcolsep}{2.2pt}
\renewcommand{\arraystretch}{1.05}
\resizebox{\textwidth}{!}{%
\begin{tabular}{c|ccc|ccccc|ccc}
\toprule
& \multicolumn{3}{c|}{Sequence Info}
& \multicolumn{5}{c|}{Same front-end: five sessions $+$ MR.ScaleMaster / single session}
& \multicolumn{3}{c}{Single-session long-context references} \\
\cmidrule(lr){2-4}\cmidrule(lr){5-9}\cmidrule(lr){10-12}
Seq.
& $L$ [m]
& Dur. [s]
& KF
& CUT3R~\cite{wang2025cut3r}
& TTT3R~\cite{chen2026ttt3r}
& \makecell{VGGT-SLAM\\2.0~\cite{maggio2025vggt-slam2}}
& LoGeR~\cite{zhang2026loger}
& LingBot-Map~\cite{chen2026lingbot}
& VGGT-Long~\cite{deng2025vggt}
& SwiftVGGT~\cite{lee2026swiftvggt}
& VGGT-Align~\cite{zhang2026vggt} \\
\midrule
0
& 617.7 & 821.3 & 579
& \textbf{35.07} / 77.95
& \textbf{11.96} / 15.19
& 26.37 / \textbf{15.31}
& \textbf{2.32} / 9.72
& \textbf{3.30} / 11.39
& 41.04 & 68.35 & 65.93 \\
3
& 1004.7 & 1472.8 & 1180
& \textbf{54.92} / 71.46
& \textbf{10.91} / 37.61
& \textbf{30.23} / 34.83
& \textbf{8.43} / 22.96
& \textbf{8.41} / 27.71
& 54.75 & 71.32 & 65.75 \\
4
& 620.8 & 823.7 & 361
& \textbf{41.26} / 58.37
& \textbf{39.94} / 65.12
& 34.69 / \textbf{18.97}
& \textbf{2.34} / 31.93
& \textbf{7.31} / 17.67
& 72.56 & 72.16 & 71.82 \\
6
& 1142.2 & 1358.2 & 1663
& \textbf{52.88} / 110.72
& \textbf{29.50} / 47.73
& \textbf{35.89} / 121.57
& \textbf{6.35} / 59.73
& \textbf{8.04} / 55.91
& 86.84 & 105.77 & 119.08 \\
16
& 1404.5 & 1706.5 & 2325
& \textbf{102.87} / 125.00
& \textbf{50.82} / 103.22
& \textbf{46.11} / 144.57
& \textbf{11.54} / 38.67
& \textbf{13.66} / 77.64
& 100.73 & 129.56 & 142.28 \\
21
& 1644.4 & 2153.1 & 2097
& 136.47 / \textbf{113.67}
& \textbf{58.82} / 65.80
& \textbf{78.14} / 135.33
& \textbf{16.00} / 47.90
& \textbf{10.83} / 98.07
& 111.81 & 130.46 & 112.53 \\
22
& 1175.7 & 1551.6 & 1404
& \textbf{81.79} / 112.04
& \textbf{36.39} / 55.45
& \textbf{33.88} / 117.78
& \textbf{7.09} / 60.62
& \textbf{6.42} / 61.75
& 114.57 & 103.66 & 106.54 \\
\midrule
Mean
& -- & -- & --
& \textbf{72.18} / 95.60
& \textbf{34.05} / 55.73
& \textbf{40.76} / 84.05
& \textbf{7.72} / 38.79
& \textbf{8.28} / 50.02
& 83.19 & 97.33 & 97.70 \\
\bottomrule
\end{tabular}}
\vspace{-4mm}
\end{table*}

%% file: src_v3/tables/tab_scale_alarm_full_stack.tex
\begin{table}[t!]
\centering
\caption{\acs{SCA} on 22 ScaleMaster segments~\cite{ju2026have}. ATE RMSE
[m] is reported without (OFF) and with (ON) the alarm;
$\Delta=\mathrm{ON}-\mathrm{OFF}$. \emph{Straight}/\emph{Rollback} test the
two gates, and \emph{TP} contains genuine revisits. DIV denotes divergence;
shading marks DIV recovery or $\Delta\leq-4$\,m.}
\label{tab:exp_alarm}
\vspace{-2.0mm}
\footnotesize
\setlength{\tabcolsep}{4pt}
\renewcommand{\arraystretch}{0.9}
\resizebox{0.85\columnwidth}{!}{%
\begin{tabular}{@{}llrrr@{}}
\toprule
Cat. & Segment & OFF & ON & $\Delta$ \\
\midrule
\multirow{7}{*}{Straight}
 & S1 LargeHall\_01 (14600--18600)  & 5.56  & \textbf{3.61}  & $-1.95$ \\
 & \cellcolor{green!12}S2 LargeHall\_02 (1400--5400) & \cellcolor{green!12}DIV & \cellcolor{green!12}\textbf{8.44} & \cellcolor{green!12}-- \\
 & S3 Parking\_01 (5000--7500)      & 2.07  & \textbf{1.66}  & $-0.41$ \\
 & \cellcolor{green!12}S4 Parking\_01 (800--3300) & \cellcolor{green!12}23.03 & \cellcolor{green!12}\textbf{0.39} & \cellcolor{green!12}$-22.64$ \\
 & \cellcolor{green!12}S5 Stairs\_01 (3200--4700) & \cellcolor{green!12}11.72 & \cellcolor{green!12}\textbf{0.43} & \cellcolor{green!12}$-11.29$ \\
 & S6 Library\_01 (3400--4900)      & 0.32  & \textbf{0.27}  & $-0.05$ \\
 & S7 Library\_02 (600--2100)       & \textbf{0.18}  & 0.19  & $+0.01$ \\
\cmidrule(l){1-5}
\multirow{7}{*}{Rollback}
 & \cellcolor{green!12}RB1 Library\_01 (1000--2500) & \cellcolor{green!12}4.61 & \cellcolor{green!12}\textbf{0.46} & \cellcolor{green!12}$-4.15$ \\
 & RB2 Library\_02 (1400--4900)     & 16.77 & \textbf{15.16} & $-1.61$ \\
 & RB3 Library\_02 (600--3600)      & 7.89  & \textbf{7.29}  & $-0.60$ \\
 & RB4 Library\_03 (100--2800)      & 5.18  & \textbf{1.60}  & $-3.58$ \\
 & RB5 LargeHall\_04 (0--4000)      & \textbf{0.35}  & 0.61  & $+0.26$ \\
 & RB6 Parking\_01 (2000--5000)     & \textbf{9.34}  & 9.72  & $+0.38$ \\
 & RB7 Library\_04 (2400--5400)     & 4.04  & \textbf{2.94}  & $-1.10$ \\
\cmidrule(l){1-5}
\multirow{8}{*}{TP}
 & TP1 Lobby\_01 (full)             & 0.95  & \textbf{0.66}  & $-0.29$ \\
 & \cellcolor{green!12}TP2 Library\_02 (full) & \cellcolor{green!12}11.04 & \cellcolor{green!12}\textbf{1.58} & \cellcolor{green!12}$-9.46$ \\
 & \cellcolor{green!12}TP3 Library\_03 (full) & \cellcolor{green!12}15.47 & \cellcolor{green!12}\textbf{0.83} & \cellcolor{green!12}$-14.64$ \\
 & TP4 Stairs\_02 (full)            & \textbf{1.49}  & 2.46  & $+0.97$ \\
 & TP5 Lounge\_01 (full)            & 2.00  & \textbf{1.31}  & $-0.69$ \\
 & TP6 LargeHall\_04 (full)         & \textbf{0.45}  & 1.00  & $+0.55$ \\
 & TP7 Parking\_02 (full)           & \textbf{0.62}  & 2.93  & $+2.31$ \\
 & TP8 HotelRoom\_01 (1200--4050)   & 0.08  & 0.08  & $0.00$ \\
\bottomrule
\end{tabular}}
\vspace{-5mm}

\end{table}

%% file: src_v3/figures/fig_coda_multi_robot_bev.tex
\begin{figure}[t!]
    \centering
    \includegraphics[width=1.0\columnwidth]{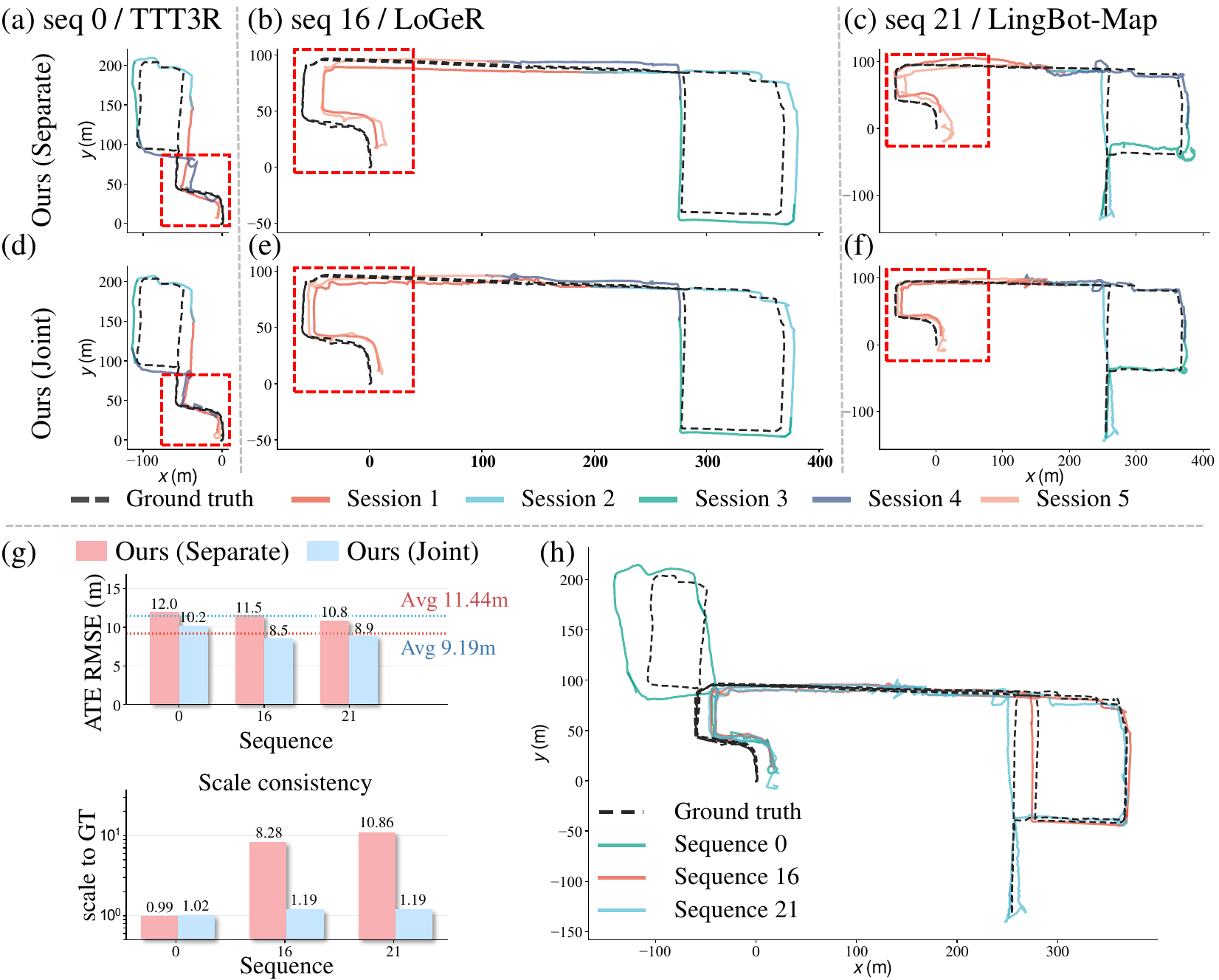}
    \caption{Joint fusion of \texttt{CODa~0} (TTT3R), \texttt{CODa~16}
    (LoGeR), and \texttt{CODa~21} (LingBot-Map), five sessions each. (a)--(c)
    Separate fusion; (d)--(f) joint
    15-session fusion, with red boxes marking reduced distortion. (g)
    Per-sequence ATE and ground-truth alignment scale in meters per reconstructed
    unit; scale spread indicates cross-front-end gauge consistency. (h) Joint
    trajectories overlaid on ground truth.}
    \label{fig:coda_multi_robot_bev}
    \vspace{-5mm}
\end{figure}

%% file: src_v3/limitation.tex
\section{Limitations}
\label{sec:limitations}

MR.ScaleMaster currently requires retrievable visual overlap between sessions;
large reverse-viewpoint changes can therefore leave sessions disconnected.
Beyond correspondence verification and a Huber kernel, inter-session loop
factors are not screened for false positives; a spurious factor that passes
verification enters the graph and, as noted in \secref{sec:rel}, can be
absorbed through scale distortion rather than exposed by its residual.
The \acs{SCA} applies only to front-ends that expose an incremental scale
trajectory and support factor rollback. These limitations motivate
viewpoint-robust retrieval, scale-aware outlier rejection for inter-session
loops, and broader alarm interfaces.

%% file: src_v3/conclusion.tex
\section{Conclusion}
\label{sec:conclusion}

We presented MR.ScaleMaster, a cooperative mapping backend for independently
selected monocular reconstruction front-ends. It constructs cross-session
$\mathrm{Sim}(3)$ \acs{CFL}s from input packet geometry. \acs{SPC} initializes
the anchor-node graph, which then corrects residual scale and drift. On KITTI
with ground-truth-derived path-length scale ratios of
$57$--$103\times$, the full system reduces mean ATE from 151.23 to
39.82\,m relative to RayOpt loop factors. On CODa, it lowers mean ATE for all
five front-ends and jointly registers 15 sessions from three front-ends in one
map. On 22 ScaleMaster segments, the agent-side \acs{SCA} recovers the only
divergent run, and a four-agent deployment demonstrates real-world
heterogeneous mapping. To our knowledge, this is the first system to fuse
monocular sessions from independently selected front-ends with incompatible
scale conventions.